\documentclass[10pt,twocolumn,letterpaper]{article}

\usepackage[pagenumbers]{cvpr} % To force page numbers, e.g. for an arXiv version

\usepackage[dvipsnames]{xcolor}

\definecolor{cvprblue}{rgb}{0.21,0.49,0.74}
\usepackage[pagebackref,breaklinks,colorlinks,citecolor=cvprblue]{hyperref}
 
\usepackage{amsmath}
\usepackage{amssymb}
\usepackage{xcolor}
\definecolor{Graylight}{gray}{0.9}
\definecolor{Gray}{gray}{1.0}
\usepackage{colortbl}
\usepackage{booktabs}
\usepackage{multirow}
\usepackage{mathrsfs}
\usepackage{marvosym}
\usepackage{enumerate}
\usepackage[hang,flushmargin]{footmisc} 
\usepackage{tablefootnote}

\definecolor{c1}{HTML}{C9461C}
\definecolor{c2}{HTML}{249087}
\definecolor{c3}{HTML}{314272}
\definecolor{c4}{HTML}{0070C0}
\definecolor{c5}{HTML}{02B0F0}

\def\paperID{*****} % *** Enter the Paper ID here
\def\confName{CVPR}
\def\confYear{2024}

\title{InternLM-XComposer-2.5: A Versatile Large Vision Language Model Supporting Long-Contextual Input and Output}

\author{Pan Zhang$^{*1}$, Xiaoyi Dong$^{*1,2}$, Yuhang Zang$^{*1}$, Yuhang Cao$^{1}$, Rui Qian$^{1,2}$, Lin Chen$^{1}$, Qipeng Guo$^{1}$, \\ Haodong Duan$^{1}$, Bin Wang$^{1}$, Linke Ouyang$^{1}$, Songyang Zhang$^{1}$, Wenwei Zhang$^{1}$, Yining Li$^{1}$, \\ Yang Gao$^{1}$, Peng Sun$^{1}$, Xinyue Zhang$^{1}$, Wei Li$^{1}$, Jingwen Li$^{1}$, 
Wenhai Wang$^{1,2}$, Hang Yan$^{1}$, \\ Conghui He$^{3}$, Xingcheng Zhang$^{3}$, Kai Chen$^{1}$, Jifeng Dai$^{4,1}$, Yu Qiao$^{1}$, Dahua Lin$^{1,2}$, Jiaqi Wang$^{1,}${\textsuperscript{\Letter}}\\
$^1$Shanghai Artificial Intelligence Laboratory,  $^2$The Chinese University of Hong Kong, \\ $^3$SenseTime Group, $^4$Tsinghua University \\
\tt\small
internlm@pjlab.org.cn
}

\begin{document}
\maketitle
{\let\thefootnote\relax\footnotetext{\noindent* equal contribution. \Letter~corresponding author.}}

\begin{abstract}
    We present InternLM-XComposer-2.5 (IXC-2.5), a versatile large-vision language model that supports long-contextual input and output. \textbf{IXC-2.5 excels in various text-image comprehension and composition applications, achieving GPT-4V level capabilities with merely 7B LLM backend.} Trained with 24K interleaved image-text contexts, it can seamlessly extend to 96K long contexts via RoPE extrapolation. This long-context capability allows IXC-2.5 to excel in tasks requiring extensive input and output contexts. Compared to its previous 2.0 version, InternLM-XComposer-2.5 features three major upgrades in vision-language comprehension: (1) Ultra-High Resolution Understanding, (2) Fine-Grained Video Understanding, and (3) Multi-Turn Multi-Image Dialogue. In addition to comprehension, IXC-2.5 extends to two compelling applications using extra LoRA parameters for text-image composition: (1) Crafting Webpages and (2) Composing High-Quality Text-Image Articles. IXC-2.5 has been evaluated on 28 benchmarks, outperforming existing open-source state-of-the-art models on 16 benchmarks. It also surpasses or competes closely with GPT-4V and Gemini Pro on 16 key tasks. The InternLM-XComposer-2.5 is publicly available at \url{https://github.com/InternLM/InternLM-XComposer}. 
\end{abstract}    
\section{Introduction}\label{sec:intro}

\begin{figure}[t!]
    \centering
    %\vspace{-20pt}
    \includegraphics[width=1.0\linewidth]{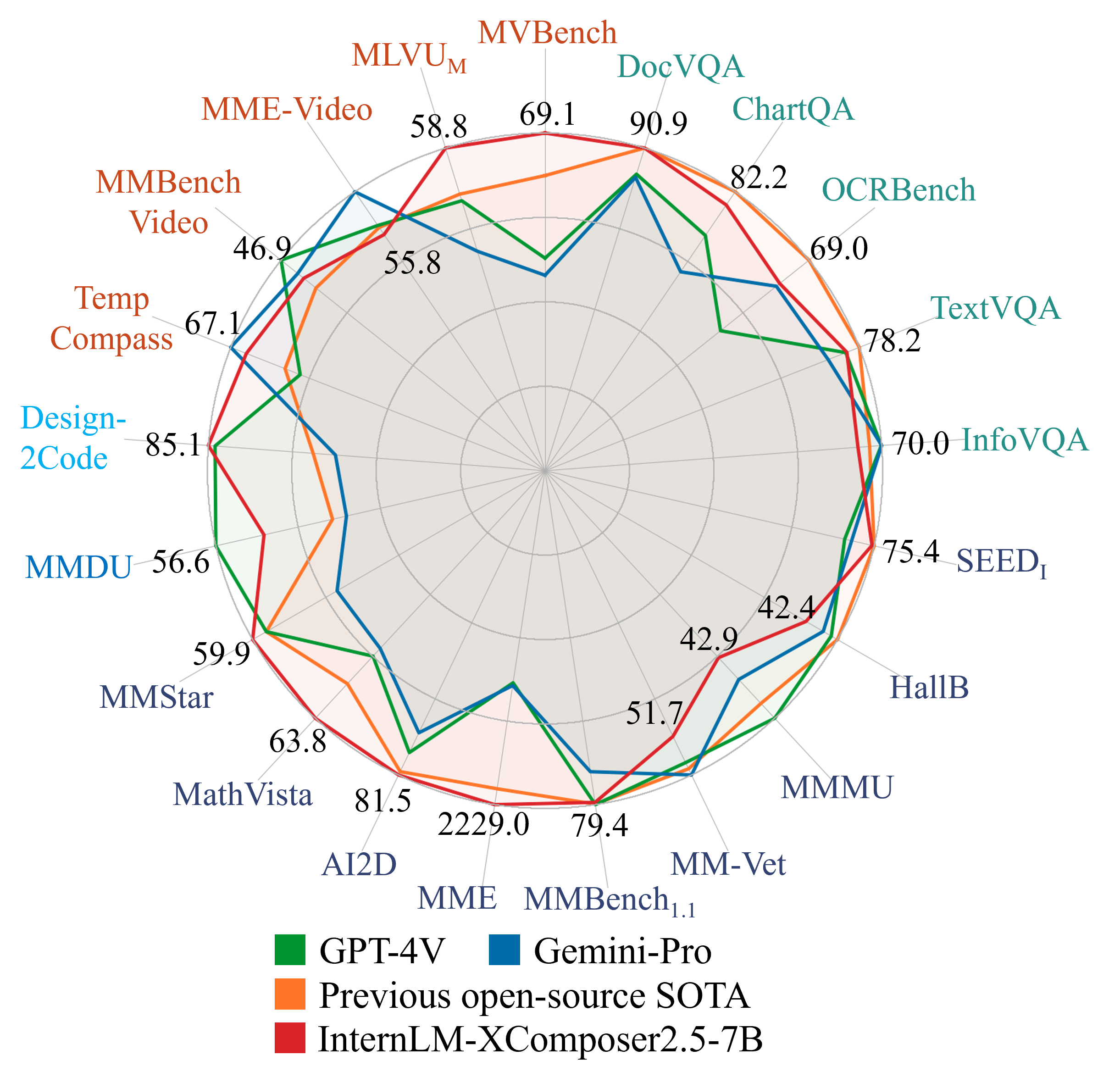}
    
    \setlength{\abovecaptionskip}{0mm} 
    \captionof{figure}{\small
        Overview of InternLM-XComposer-2.5 (IXC-2.5) performance on benchmarks in different domains, including \textbf{{\color{c1}{Video Benchmarks}}}, \textbf{{\color{c2}{Structural High-resolution Benchmarks}}}, \textbf{{\color{c3}{General Visual QA Benchmarks}}}, \textbf{{\color{c4}{Multi-True Multi-Image Benchmark}}}, and \textbf{{\color{c5}{Webpage Crafting Benchmark}}}.
        IXC-2.5 based on InternLM2-7B~\cite{2023internlm} \textbf{matches or even surpasses GPT-4V~\cite{openai2023gpt4} and Gemini Pro~\cite{geminiteam2023gemini} in 15 benchmarks. Please refer to Table~\ref{tab:sota_comp_1},~\ref{tab:sota_comp_2},~\ref{tab:sota_comp_design2code} for details.}
	}
	\label{fig:teaser}
    %\vspace{-5pt}
\end{figure}

Recent advancements in Large Language Models (LLMs)~\cite{openai2020chatgpt,chowdhery2022palm,touvron2023llama,touvron2023llama2,jiang2023mistral,qwen7b} have sparked interest in the development of Large Vision Language Models (LVLMs)~\cite{openai2023gpt4,fu2023gemini,zhu2023minigpt,dai2023instructblip,zhang2023internlm,liu2023visual}. Leading paradigms like GPT-4~\cite{openai2023gpt4}, Gemini Pro 1.5~\cite{fu2023gemini}, and Claude 3~\cite{Claude_3} have achieved considerable success and significantly expanded the range of applications for LLMs. Open-source LVLMs are also being rapidly developed and can compete with proprietary APIs in several benchmarks. However, these open-source models still lag behind closed-source leading paradigms in versatility. They lack the ability to perform diverse vision-language comprehension and composition tasks, largely due to limited diversity in training corpus and challenges in managing long-context input and output.

To further bridge the gap between proprietary APIs~\cite{openai2023gpt4,fu2023gemini} and open-sourced Large Vision Language Models, we are introducing InternLM-XComposer-2.5 (IXC-2.5), a versatile LVLM supporting long-contextual input and output with diverse comprehension and composition capacities. IXC-2.5 excels in existing open-sourced LVLMs with two advantages. \textbf{(1) Versatility}: IXC-2.5 supports a wide range of tasks related to comprehension and composition, such as free-form text-image conversation, OCR, video understanding, article composition with illustrations, and webpage crafting. \textbf{(2) Long-context capabilities in both input and output:} It is natively trained with 24K interleaved image-text data, whose context window can be extended to 96K through positional encoding extrapolation~\cite{dynamicNTK}, empowering the long-term human-AI interaction and content creation. 

Benefiting from the long contextual capability, compared to its previous 2.0 version~\cite{internlmxcomposer2}, IXC-2.5 has upgraded three comprehension abilities: \textbf{(1) Ultra-High Resolution Understanding:} IXC-2.5 enhances the dynamic resolution solution proposed in IXC2-4KHD~\cite{internlmxcomposer2_4khd} with a native 560 × 560 ViT vision encoder, supporting high-resolution images with any aspect ratio. \textbf{(2) Fine-Grained Video Understanding:} IXC-2.5 treats videos as a ultra-high-resolution composite picture consisting of tens to hundreds of frames, allowing it to capture fine details through dense sampling and higher resolution for each frame. \textbf{(3) Multi-Turn Multi-Image Dialogue:} IXC-2.5 supports free-form multi-turn multi-image dialogue, allowing it to naturally interact with humans in multi-round conversations. 

Besides comprehension, IXC-2.5 also supports two notable applications by incorporating extra LoRA parameters for text-image composition:
\textbf{(1) Crafting Webpages:} IXC-2.5 can be readily applied to create webpages by composing source code (HTML, CSS, and JavaScript) following text-image instructions. \textbf{(2) Composing High-Quality Text-Image Articles:} Compared to IXC-2, IXC-2.5 leverages specially designed Chain-of-Thought (CoT)~\cite{wei2023chainofthoughtpromptingelicitsreasoning} and Direct Preference Optimization (DPO)~\cite{rafailov2024direct} techniques to significantly enhance the quality of its written content. 

We evaluated the versatility of InternLM-XComposer-2.5 (IXC-2.5) across a range of twenty-eight benchmarks, including five video benchmarks~\cite{li2024mvbench,zhou2024mlvu,fu2024video,fang2024mmbench,liu2024tempcompass}, nine structural high-resolution benchmarks~\cite{docvqa,masry2022chartqa,infovqa,textvqa,ocrbench,deepform,pasupat-liang-2015-compositional,tanaka2021visualmrc,chen2020tabfact}, twelve general VQA benchmarks~\cite{mmstar,lu2024mathvista,yue2023mmmu,kembhavi2016diagram,fu2023mme,MMBench,li2023seedbench,yu2023mmvet,guan2023hallusionbench,RealWorldQA}, one multi-true multi-image benchmark~\cite{liu2024mmdu}, and one webpage crafting benchmark~\cite{si2024design2code}.
Compared to previous open-source LVLMs, IXC-2.5 achieved state-of-the-art results in 16 out of 28 benchmarks based on InternLM2-7B~\cite{2023internlm} backend. As shown in Figure~\ref{fig:teaser}, the performance of IXC-2.5 matches or even surpasses proprietary APIs, \eg, GPT-4V~\cite{openai2023gpt4} and Gemini Pro ~\cite{fu2023gemini}, in 16 benchmarks.

IXC-2.5 web demo now supports audio input and output using open-source tools~\cite{radford2022robust,zhao2024melo}. You may try it at \url{https://huggingface.co/spaces/Willow123/InternLM-XComposer}.

\begin{figure*}[h!]
    \centering
    \vspace{-40pt}
    \includegraphics[width=0.97\linewidth]{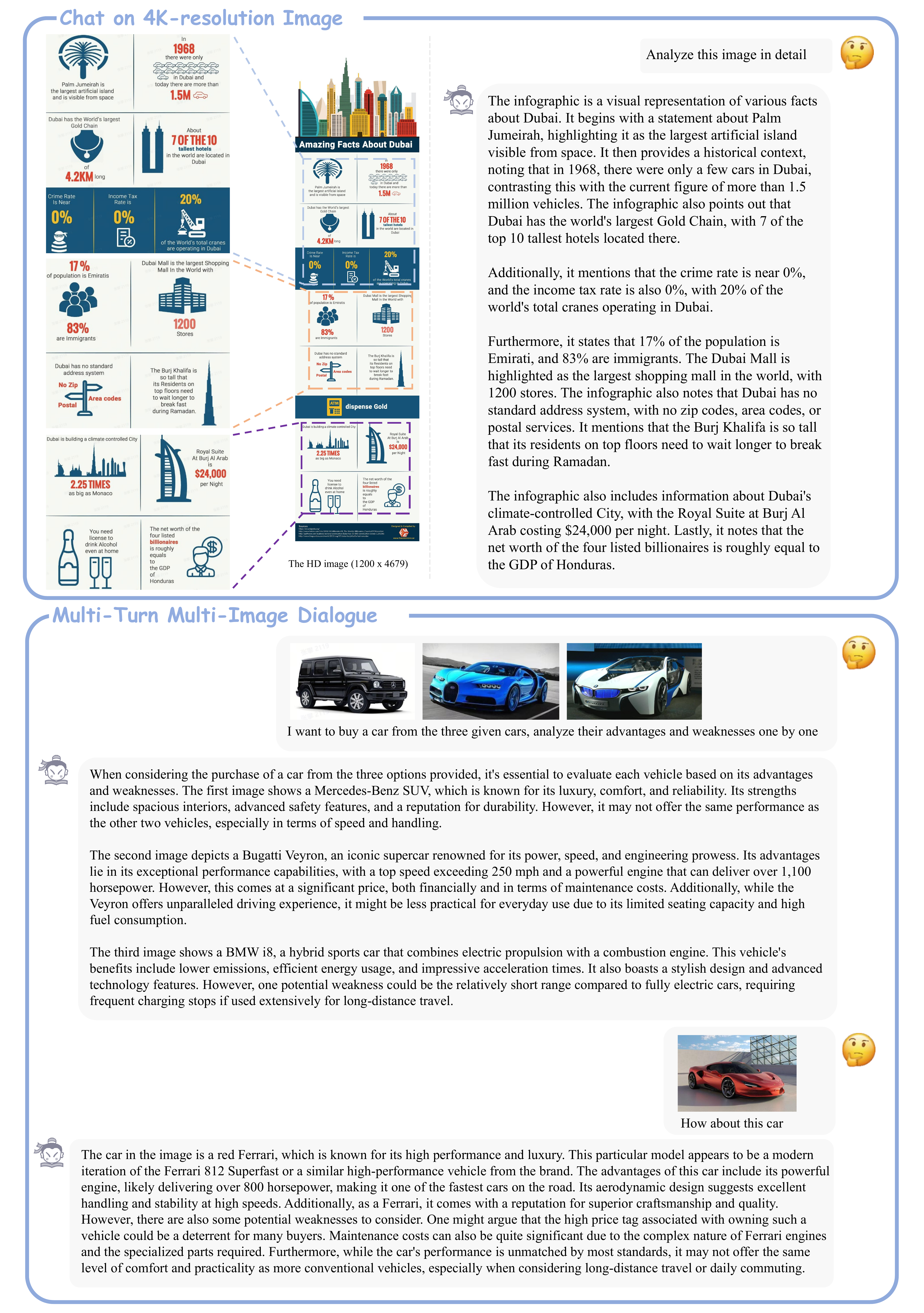}
    
    \setlength{\abovecaptionskip}{0mm} 
    \captionof{figure}{\small
        The capabilities of IXC-2.5: Chat on 4K-resolution Image and Multi-Turn Multi-Image Dialogue.
	}
	\label{fig:teaser1}
    \vspace{-20pt}
\end{figure*}

\begin{figure*}[h!]
    \centering
    \vspace{-40pt}
    \includegraphics[width=0.97\linewidth]{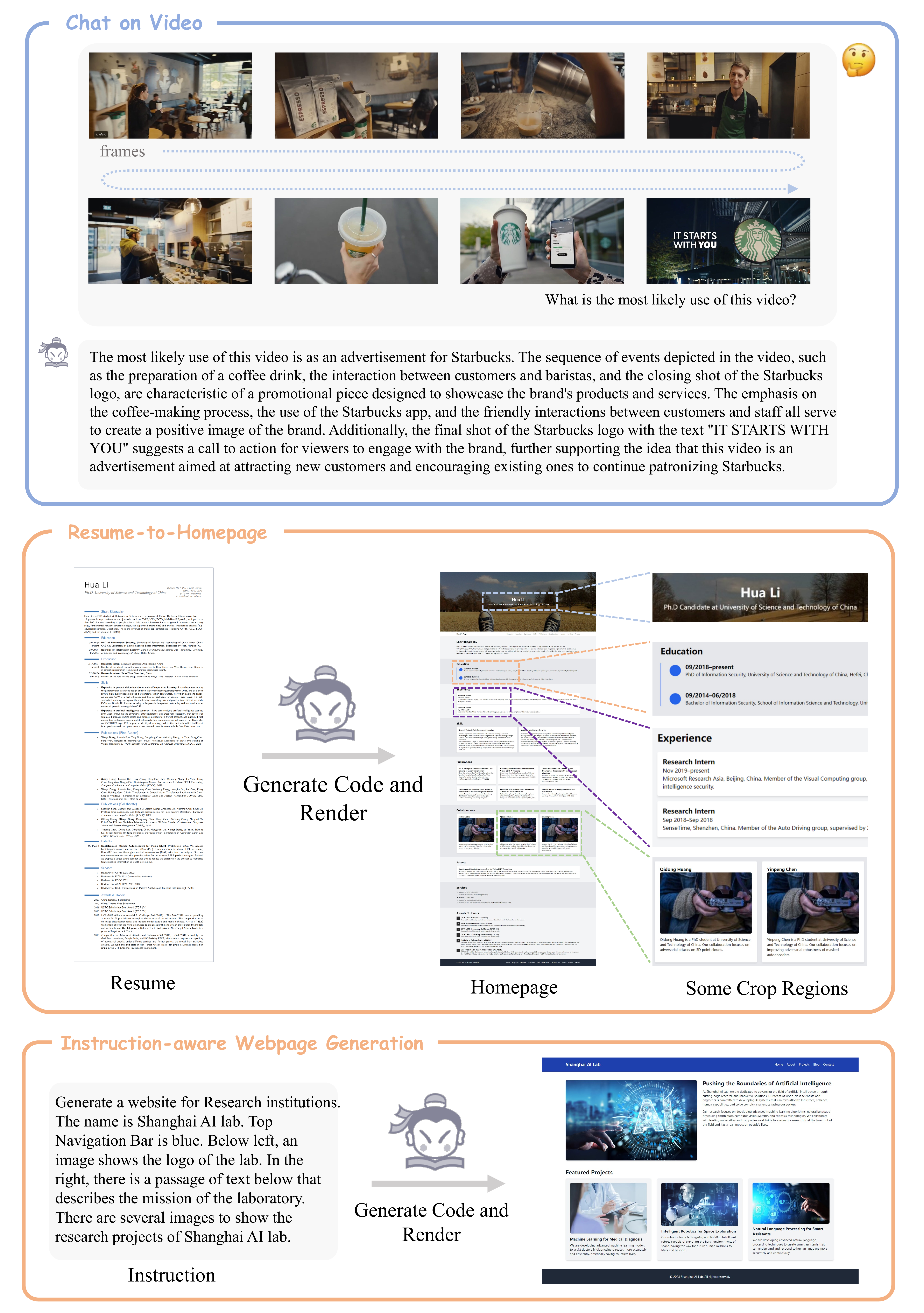}
    
    \setlength{\abovecaptionskip}{0mm} 
    \captionof{figure}{\small
        The capabilities of IXC-2.5: Chat on Video, Resume-to-Homepage and Instruction-aware Webpage Generation.
	}
	\label{fig:teaser2}
    \vspace{-20pt}
\end{figure*}

\begin{figure*}[h!]
    \centering
    \vspace{-40pt}
    \includegraphics[width=0.97\linewidth]{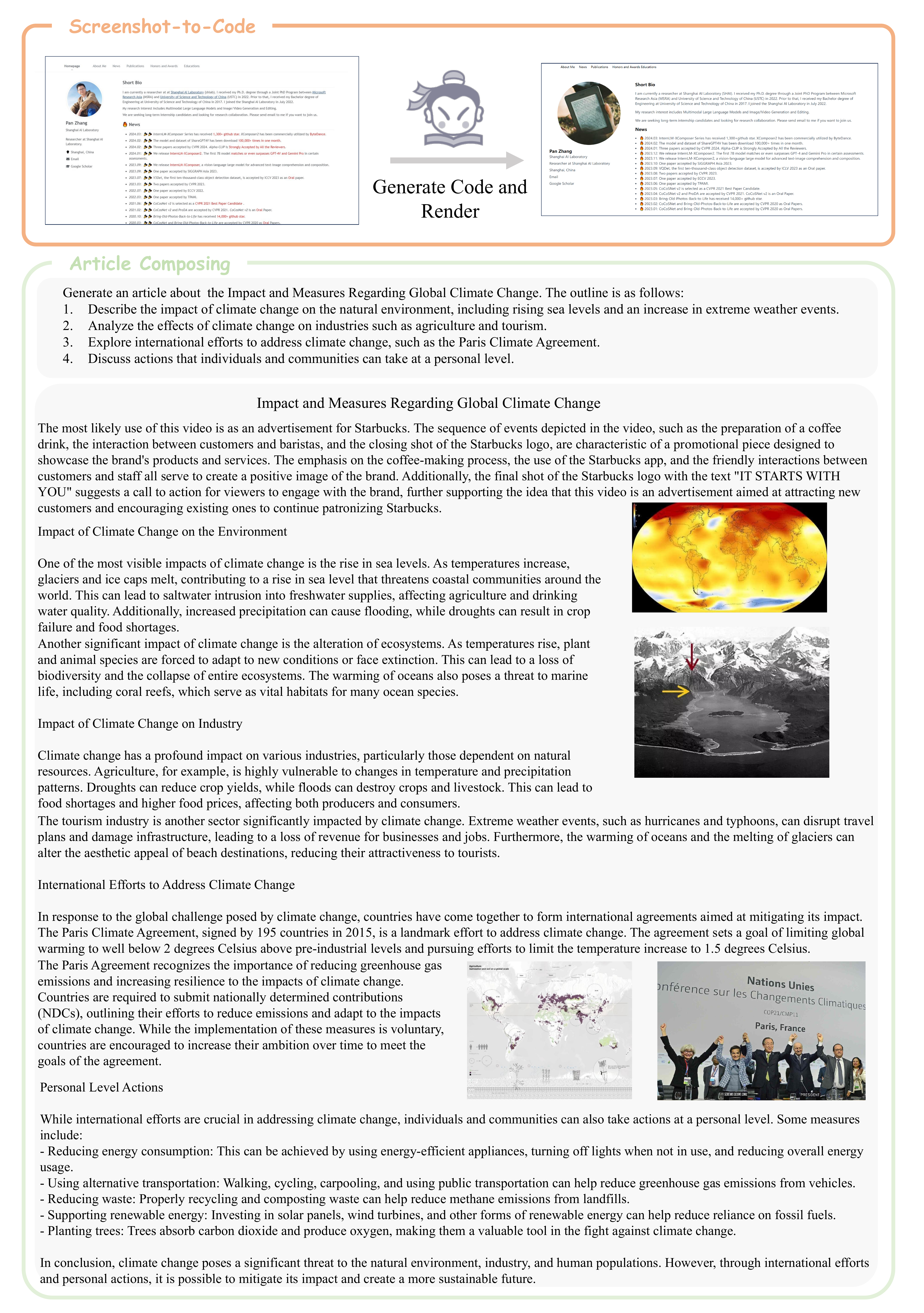}
    
    \setlength{\abovecaptionskip}{0mm} 
    \captionof{figure}{\small
        The capabilities of IXC-2.5: Screenshot-to-Code and Article Composing.
	}
	\label{fig:teaser3}
    \vspace{-20pt}
\end{figure*}
\section{Related Works}\label{sec:related_works}

\noindent \textbf{LVLMs for Text-Image Conversation.} Large Language Models (LLMs)~\cite{brown2020language,ouyang2022training,openai2020chatgpt,chowdhery2022palm,kaplan2020scaling,touvron2023llama,touvron2023llama2,jiang2023mistral,2023internlm,zeng2023glm-130b,baichuan2023baichuan2,qwen7b,cai2024internlm2} have received considerable attention because of their impressive performance in language comprehension and generation. Large vision-language models (LVLMs)~\cite{openai2023gpt4,chen2023pali,chen2023palix,chen2023pali3,driess2023palme,fu2023gemini,zhu2023minigpt,dai2023instructblip,zhang2023internlm,fuyu-8b,li2023otter,peng2023kosmos,ye2023mplug,awadalla2023openflamingo,internlmxcomposer2_4khd,lin2024vilapretrainingvisuallanguage} have been developed by integrating LLMs with vision encoders~\cite{radford2021learning,zhang2024long,sun2023alpha,zhai2023sigmoid,oquab2024dinov2learningrobustvisual,zang2023contextual,liu2022convnet2020s,chen2023internvl,chen2023sharegpt4v,lin2023sphinx,bai2023qwen,wang2023cogvlm,internlmxcomposer2,cao2024dualfocus,liu2024rar,chen2024fargpt4vclosinggap,zhang2024omgllavabridgingimagelevelobjectlevel} to extend the ability to understand vision content, enabling the application of text-image conversation. Most existing LVLMs are trained for single-image multi-round conversations, while some works~\cite{alayrac2022flamingo,bai2023qwen,zhao2023mmicl,sun2024generativemultimodalmodelsincontext,lin2024vilapretrainingvisuallanguage,jiang2024mantisinterleavedmultiimageinstruction} have the ability to understand multi-image inputs. However, IXC-2.5 focuses on providing a free-form long-contextual multi-turn multi-image interaction experience~\cite{liu2024mmdu,liu2024convbench,ma2024mmlongbench}, which has not been addressed yet.

\noindent \textbf{LVLMs for High Resolution Images Analysis.} Understanding high-resolution images has significant potential applications such as OCR and document/chart analysis, which is attracting increased attention in the LVLMs area. In recent works, there are two main strategies to enable high-resolution understanding: (1) High-resolution (HR) visual encoders~\cite{lv2023kosmos25,wei2023vary,cogagent,li2024mini,tang2024textsquarescalingtextcentricvisual,zhang2024llavahddivinghighresolutionlarge} directly support higher resolution images. (2) Patchification: A high-resolution image is cropped into patches~\cite{li2023monkey,monkeytext,llavauhd,ureader,docowl,lin2023sphinx,li2023otterhd,lin2023sphinx,llavanext,ureader,internlmxcomposer2_4khd}. Each patch is processed with a low resolution vision encoder, \eg, CLIP~\cite{radford2021learning} and visual embeddings of patches are further concatenated as inputs for LLM backends. IXC-4KHD~\cite{internlmxcomposer2_4khd} scales the supported resolution of open-source LVLMs into 4K and beyond for the first time. IXC-2.5 combines both solutions with a vision encoder trained with a resolution of 560x560 and a dynamic resolution solution proposed in IXC2-4KHD~\cite{internlmxcomposer2_4khd}, resulting in further improvements.

\noindent \textbf{LVLMs for Video Understanding.} In addition to image understanding, the LVLMs area has also witnessed emerging efforts in video analysis~\cite{li2023mvbench,ning2023video, liu2024tempcompass, caba2015activitynet,fang2024mmbenchvideolongformmultishotbenchmark,song2023moviechat,song2024moviechat+}. To handle complex video inputs, existing works use sparse sampling or temporal pooling~\cite{lin2023video, maaz2023video,luo2023valley,huang2024image,yu2024self}, compressed video tokens~\cite{li2023videochat,zhang2023video,jin2023chat,weng2024longvlm,li2023llama,qian2024streaminglongvideounderstanding}, memory banks~\cite{song2023moviechat,he2024ma,song2024moviechat+}, and language as a bridge~\cite{kahatapitiya2024language,islam2024video,zhang2023simple} for video understanding. Apart from these video-specific designs, video analysis can also be formulated to understand a high-resolution composite picture consisting of sampled video frames~\cite{kim2024imagegridworthvideo,xu2024pllavaparameterfreellava,zhang2024longcontexttransferlanguage}. Benefiting from the ability to comprehend ultra-high-resolution images and long context, IXC-2.5 exhibits strong performance on various video benchmarks for LVLMs.

\noindent \textbf{Webpage Generation.} Pix2Code~\cite{beltramelli2018pix2code} presents an end-to-end solution for UI-to-code transformation leveraging CNNs and RNNs. This approach contends with the challenges posed by intricate visual encoding and extensive text decoding when applied to real-world UIs. In the sphere of recent advancements, works such as Sightseer~\cite{laurenccon2024unlocking}, DCGen~\cite{wan2024automatically}, and Design2Code~\cite{si2024design2code} have employed large vision-language models trained on synthetic screenshot-HTML paired datasets like WebSight v0.1 or v0.2~\cite{laurenccon2024unlocking} to facilitate HTML code generation. Nevertheless, the synthesized web page datasets have been critiqued for their simplicity and lack of diversity. These studies generally concentrate on the screenshot/sketch-to-code task. In contrast, our IXC-2.5 model extends these capabilities to include screenshot-to-code, instruction-aware webpage generation, and resume-to-homepage tasks. IXC-2.5 is trained using a combination of high-quality synthesized and real-world web data. Furthermore, IXC-2.5 is proficient in generating JavaScript code, thereby enabling the development of interactive front-end webpages.

\noindent \textbf{Preference Alignment.} Reinforcement Learning from Human Feedback (RLHF)~\cite{ouyang2022training} and Reinforcement Learning from AI Feedback (RLAIF)~\cite{bai2022constitutional} have shown great promise in aligning LLMs across various domains, including improving logical reasoning and generating helpful and harmless outputs. The typical approach involves training a reward model using human or AI preference data and fine-tuning the LLM to maximize the expected reward function with optimization algorithms like Proximal Policy Optimization (PPO)~\cite{schulman2017proximal}. Alternatively, Direct Preference Optimization (DPO)~\cite{rafailov2024direct} and the following works~\cite{pal2024smaug,ethayarajh2024kto} have emerged as leading methods that implicitly represent the reward score and eliminate the need for a separate reward model. Building on the success of RLHF and RLAIF in LLMs, recent studies have successfully extended RLHF/RLAIF algorithms for multimodal LVLMs~\cite{li2023silkie,yu2024rlhf,zhou2024aligning,zhao2023beyond,pi2024strengthening} to reduce hallucination. In this work, we investigate the application of preference alignment techniques to the text-image article composition task, with a focus on generating high-quality and stable response results.
\section{Method}

\begin{figure}
    \centering
    \includegraphics[width=1.\linewidth]{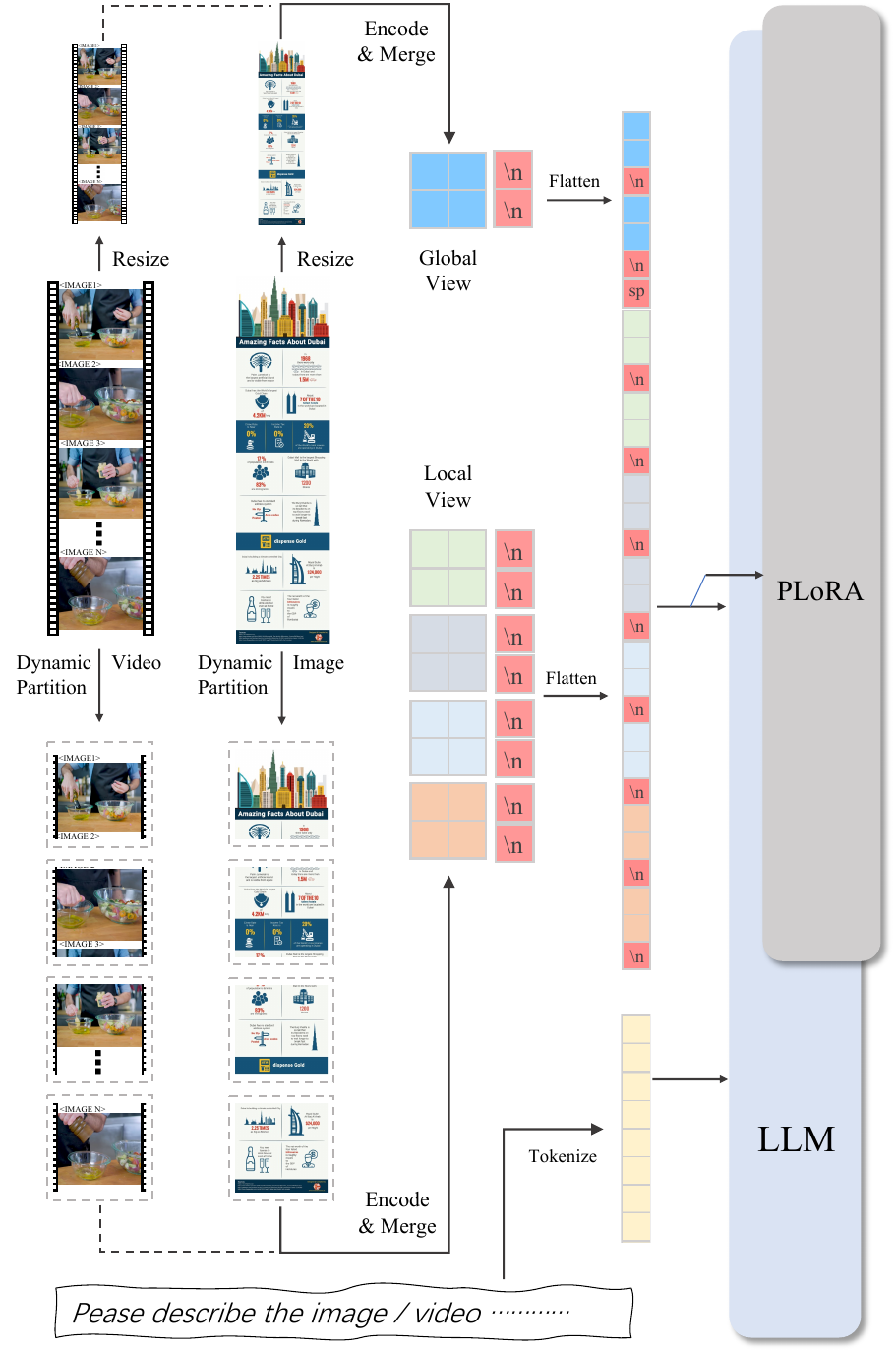}
    \caption{{\textbf{Framework}} of IXC-2.5 that supports the multi-modal inputs, including text, single/multiple images, and videos. }
    \label{fig:framework}
\end{figure}

\subsection{Model Architecture}
The model architecture of InternLM-XComposer-2.5 (IXC-2.5 in the following for simplicity) mainly follows the design of InternLM-XComposer2~\cite{internlmxcomposer2} and InternLM-XComposer2-4KHD~\cite{internlmxcomposer2_4khd} (IXC2 and IXC2-4KHD for simplicity), including a light-weight Vision Encoder OpenAI ViT-L/14~\cite{radford2021learning}, Large Language Model InternLM2-7B~\cite{cai2024internlm2}, and Partial LoRA~\cite{internlmxcomposer2} for efficient alignment. We recommend the readers to the IXC2 and IXC2-4KHD papers for more details.

\subsection{Multi-modal Input}
Our IXC-2.5 supports diverse input modalities, including text, single/multiple images, and videos. As showin in Figure~\ref{fig:framework}, a Unified Dynamic Image Partition strategy is adopted for both videos and multiple images with any resolutions and aspect ratios.

\noindent\textbf{Image Processing.}
We mainly follow the Dynamic Image Partition and Global-Local Format design used in IXC2-4KHD~\cite{internlmxcomposer2_4khd} with a few modifications.  For the vision encoder, we reuse the ViT of $490\times490$ resolution used in IXC2 and further increase its resolution to $560\times560$, so that each sub-image has 400 tokens. 

For the high-resolution strategy, we unify the different strategies used in the IXC-4KHD into a scaled identity strategy. Given a maximum partition number $\mathcal{H}$, the image $x$ with size $[h,w]$ is resized and padded to the new image $\hat{x}$ with size $[p_h \times 560, p_w \times 560 ]$. This process is subject to the following constraints:
\begin{align}
    p_{w1} &\times \lceil p_{w1} \times h / w \rceil \leq \mathcal{H}; \\
    p_{w2} &= \lceil w * s / 560 \rceil \; \\
    p_w &= min(  p_{w1}, p_{w2})\;\; ;  p_h = \lceil p_w \times h / w \rceil
\end{align} 
where $s$ is the scale factor, $p_w$ and $p_h$ represent the number of patches in each row and column, respectively. 

For multi-image input, we assign an index to each image like $<$\text{IMAGE} i$>$, $i \in \{1,2,3,\ldots\}$ and format the image and text in an interleaved format.

\noindent\textbf{Video Processing.}
We sample frames from the given video and concatenate them along the short side of the frame, leading to a high-resolution image. The frame index is also written in the image to provide the temporal relation.

\noindent\textbf{Audio Processing.}
IXC-2.5 web demo supports audio input and output using open-source tools. For audio input, we employ Whisper \cite{radford2022robust} to transcribe audio into text. For audio output, we utilize MeloTTS \cite{zhao2024melo} to convert the text back into audio.

\subsection{Pre-training}

\begin{table*}[t]
\centering
\footnotesize
\setlength{\tabcolsep}{1mm}{
\begin{tabular}{ll}
\toprule
Task &  Dataset\\
\midrule
General Semantic Alignment  &  ShareGPT4V-PT~\cite{chen2023sharegpt4v}, COCO~\cite{chen2015microsoft},  Nocaps~\cite{agrawal2019nocaps}, TextCaps~\cite{sidorov2020textcaps},  LAION~\cite{schuhmann2021laion},  SBU~\cite{Ordonez_2011_im2text}, CC 3M~\cite{sharma2018conceptual}  ALLaVA~\cite{chen2024allava}  \\
World Knowledge Alignment & Concept Data~\cite{zhang2023internlm} \\
Vision Capability Enhancement & WanJuan~\cite{He2023WanJuanAC}, Flicker\cite{young2014flicker}, MMC-Inst\cite{liu2023mmc},  RCTW-17\cite{ocr_rctw}, CTW\cite{yuan2019ctw}, LSVT\cite{ocr_lsvt}, ReCTs\cite{ocr_rects}, ArT\cite{ocr_art} \\
% & In-house Chart/Doc/Info-to-Markdown \\
\bottomrule
\end{tabular}}
\vspace{-6pt}
\caption {\textbf{Datasets used for Pre-Training}. The data are collected from diverse sources for the three objectives. }
\label{tab:pretrain_data}
\vspace{-12pt}
\end{table*}

During the pre-training phase, the LLM (InternLM2-7B~\cite{2023internlm}) is frozen while both the vision encoder and Partial LoRA~\cite{internlmxcomposer2} are fine-tuned to align the visual tokens with the LLM. 
The data used for pre-training is shown in Table~\ref{tab:pretrain_data}.

In practice, we employ the CLIP ViT-L-14-490~\cite{radford2021learning} from IXC2 as the vision encoder and further increase its resolution to $560\times560$. For the Unified Dynamic Image Partition strategy~\cite{internlmxcomposer2_4khd}, we set the maximum number $\mathcal{H} = 12$ for the pertaining. 
For the Partial LoRA~\cite{internlmxcomposer2}, we set a rank of $256$ for all the linear layers in the LLM decoder block. 
Our training process involves a batch size of 4096 and spans across 2 epochs. The learning rate linearly increases to $2 \times 10^{-4}$ within the first $1\%$ of the training steps. Following this, it decreases to $0$ according to a cosine decay strategy. 
To preserve the original knowledge of the vision encoder, we apply a layer-wise learning rate (LLDR) decay strategy~\cite{internlmxcomposer2}, and the decay factor is set to $0.90$. 

\subsection{Supervised Fine-tuning}\label{sec:sft}

\begin{table}[t]
\centering
\footnotesize
\setlength{\tabcolsep}{1mm}{
\begin{tabular}{ll}
\toprule
Task &  Dataset\\
\midrule 
Caption  &  ShareGPT4V~\cite{chen2023sharegpt4v}, COCO~\cite{chen2015microsoft}, Nocaps~\cite{agrawal2019nocaps} \\\midrule
General QA  & VQAv2~\cite{VQAv2}, GQA~\cite{hudson2018gqa}, OK-VQA~\cite{marino2019ok} \\
            & VD \cite{visdial}, RD \cite{chen2023shikra}, VSR \cite{Liu2022VisualSR}, ALLaVA-QA \cite{chen2024allava} \\ \midrule
Multi-Turn QA & MMDU \cite{liu2024mmdu} \\\midrule
Science QA  & AI2D~\cite{kembhavi2016diagram}, SQA~\cite{lu2022learn}, TQA \cite{tqa}, IconQA \cite{lu2021iconqa}\\\midrule
Chart QA    & DVQA~\cite{kafle2018dvqa}, ChartQA~\cite{masry2022chartqa}, ChartQA-AUG~\cite{masry2022chartqa} \\\midrule
Math QA     & MathQA~\cite{yu2023metamath}, Geometry3K \cite{Geometry3K}, TabMWP \cite{tabmwp}, \\
&    CLEVR-MATH \cite{clevr_math}, Super \cite{clevr_super} \\\midrule
World Knowledge QA & A-OKVQA~\cite{schwenk2022okvqa}, KVQA~\cite{shah2019kvqa}, ViQuAE~\cite{lerner2022viquae} \\\midrule
OCR QA & TextVQA \cite{textvqa}, OCR-VQA \cite{ocr_vqa}, ST-VQA \cite{stvqa} \\\midrule
HD-OCR QA & InfoVQA\cite{infovqa},  DocVQA~\cite{docvqa}, TabFact~\cite{chen2020tabfact}, \\
&  WTQ~\cite{pasupat-liang-2015-compositional}, DeepForm~\cite{deepform}, Visual MRC~\cite{tanaka2021visualmrc}  \\\midrule
Video & ShareGPT4Video \cite{chen2024sharegpt4video}, ActivityNet~\cite{caba2015activitynet} \\\midrule
Conversation & LLaVA-150k~\cite{liu2023visual}, LVIS-Instruct4V~\cite{wang2023see} \\
& ShareGPT-en\&zh~\cite{vicuna2023}, InternLM-Chat~\cite{2023internlm} \\ 
\bottomrule
\end{tabular}}
\vspace{-6pt}
\caption {\textbf{Datasets used for Supervised Fine-Tuning}. We collect data from diverse sources to empower the model with different capabilities. }
\label{tab:sft data}
\vspace{-12pt}
\end{table}

We fine-tune the model with data listed in Table~\ref{tab:sft data}. The maximum number $\mathcal{H}$ of the Unified Dynamic Image Partition strategy is $24$ to handle extremely large images and videos. For video datasets, the IXC-2.5 is trained with large images concatenated by at most 64 frames.
The largest training context is set to a 24,000 context window size, where the MMDU~\cite{liu2024mmdu} dataset can achieve this limitation.
In practice, we jointly train all the components with a batch size of 2048 over 4000 steps. % Comment: Joint training details
Data from multiple sources are sampled in a weighted manner, with the weights based on the number of data from each source. 
The maximum learning rate is set to $5 \times 10^{-5}$, and each component has its own unique learning strategy.
For the vision encoder, we set the LLDR to $0.9$, which aligns with the pretraining strategy. 
For the LLM, we employ a fixed learning rate scale factor of $0.2$. This slows down the update of the LLM, achieving a balance between preserving its original capabilities and aligning it with vision knowledge.

\subsection{Webpage Generation}
We enhance the capabilities of the IXC-2.5 to include automated webpage generation. Specifically, the IXC-2.5 is now equipped to autonomously construct web pages, utilizing HTML, CSS, and JavaScript, based on input in the form of a visual screenshot, a set of free-form instructions, or a resume document. Current open-source general-purpose large language models frequently demonstrate suboptimal performance in generating HTML and CSS relative to their proficiency in natural language generation. To address this limitation, we propose training the screenshot-to-code task using extensive datasets from WebSight v0.1/v0.2~\cite{laurenccon2024unlocking}, and Stack v2~\cite{lozhkov2024starcoder}. Subsequently, we fine-tune the model with a smaller, meticulously crafted dataset consisting of instruction-aware webpage generation and personal page generation examples.

\noindent \textbf{Screenshot-to-code.} In addition to the WebSight~\cite{laurenccon2024unlocking} datasets, we preprocess the HTML and CSS code from the Stack v2~\cite{lozhkov2024starcoder} dataset to facilitate screenshot-to-code training. Initially, we combine the CSS and HTML code into a single file. Subsequently, we remove all comments, JavaScript code, and external links. Furthermore, we eliminate any CSS styles that are not referenced by the HTML code. We convert all files into screenshots, subsequently discarding those that did not render successfully. The remaining screenshots are then processed using the IXC2-4KHD~\cite{internlmxcomposer2_4khd} model to assess the quality of the web pages. Following the exclusion of low-quality web pages, we retained a final set of three remaining about 250,000 high-quality web pages.

We conduct training on the LoRA model utilizing the three aforementioned datasets. The LoRA rank is set to 512. The training protocol employs a batch size of 512 and is executed over a single epoch. Initially, the learning rate is incremented linearly to $1 \times 10^{-4}$ within the first $1\%$ of the training iterations. Subsequently, the learning rate decreases to $0$ following a cosine decay schedule.

\noindent \textbf{Instruction-aware Webpage Generation.} A pivotal attribute of large language models lies in their capability to adhere to human instructions. To facilitate web page generation based on freeform instructions, we propose constructing data through querying closed-source large language models. Specifically, we utilize GPT-4 to generate diverse instructions and concepts for web page creation, encompassing elements such as type, style, and layout. Subsequently, these instructions are harnessed to query Claude-3-sonnet~\cite{Claude_3} for the actual web page generation process. This approach results in 18,000 high-quality, instruction-aware samples. Additionally, we employ Tailwind CSS instead of traditional CSS, given its succinct nature.

\noindent \textbf{Resume-to-homepage.} In addition to instruction-aware webpage generation, we introduce a more practical task. Specifically, given a resume, the model is designed to generate a personal homepage. This homepage not only encapsulates the information present in the resume but also presents it with a well-structured and visually appealing format, improving both content organization and aesthetic layout. To generate corresponding datasets, we propose an idea-resume-homepage data generation pipeline. Initially, we leverage GPT-4 to produce resume ideas tailored for diverse personas, such as researchers, students, and engineers. GPT-4 is tasked with generating these resumes in markdown format based on the provided ideas. Upon obtaining the generated resumes, we then prompt Claude-3-sonnet~\cite{Claude_3} to create corresponding homepages from these resumes. To enhance the interactivity of these webpages, Claude-3-sonnet is also utilized to generate JavaScript events based on the HTML code. In total, we have constructed a dataset comprising 2,000 samples.

Upon constructing the dataset for instruction-aware webpage generation and resume-to-homepage, we subsequently fine-tuned the LoRA model for 10 epochs. All other experimental settings were maintained consistent with those employed during the screenshot-to-code training phase.

\subsection{Article Composing}
Generating high-quality text-image articles (\eg, poetry, novels, short stories, and essays) is a crucial capability for AI assistants, with various applications in daily life, including education and entertainment.
Building upon the IXC-2.5 SFT model $\pi$ in Section~\ref{sec:sft}, we enhance creative writing capabilities for generating high-quality text-image articles.
However, collecting high-quality text-image articles is a rare and expensive endeavor.
Direct fine-tuning on scarce instruction data can lead to unstable responses from LVLMs in most cases.
To overcome these challenges, we propose a scalable pipeline that integrates supervised fine-tuning, reward modeling, preference data collection, and DPO alignment for high-quality and stable article generation.

\noindent \textbf{Supervised Fine-tuning.}
We begin with the SFT model $\pi$ (Section ~\ref{sec:sft}) and a collection of 5,000 instruction tuning data samples $\mathcal{D}$ from IXC2~\cite{internlmxcomposer2}, focused on article writing.
Due to the limited scale of the instruction data, we use the SFT model to rewrite the original prompts using the Chain-of-Thought (CoT) technique~\cite{wei2022chain}, generating step-by-step prompts to supplement the instruction tuning data as augmented data $\mathcal{D}^{*}$.
We observe that the SFT model is more effective in generating long-form responses when using these augmented prompts.
We then train the initial model $\pi$ on the augmented instruction tuning data via LoRA~\cite{hu2022lora} with the rank of $256$ and get the model $\pi_{\text{ref}}$ to establish a starting point of our alignment pipeline.

\begin{table*}[t!]
\footnotesize
\centering
\setlength{\tabcolsep}{0.2mm}{
\resizebox{1.\linewidth}{!}{
\begin{tabular}{l|ccccc|ccccccccccc}
\toprule
 &  \multirow{ 2}{*}{{MVBench}} & \multirow{ 2}{*}{MLVU} & MME & MMB$^{*1}$ & Temp$^{*2}$ & Doc & Chart & Info & Text & OCR & \multirow{ 2}{*}{WTQ} & Deep & Visual & Tab  \\ 
 &    &   & Video & Video\;\;\; & Compass\;\; & VQA & QA & VQA & VQA & Bench & &  Form   & MRC & Fact  \\ 
\midrule 
Open-Source  & VideoChat & InternVL & LIVA & InternVL & Qwen-VL & InternVL & InternVL & InternVL & InternVL & GLM-4v & DocOwl & DocOwl & DocOwl & DocOwl  \\
Previous SOTA & 2-7B\cite{li2024mvbench} & 1.5-26B\cite{chen2024fargpt4vclosinggap} & 34B\cite{lin2024vilapretrainingvisuallanguage} & 1.5-26B\cite{chen2024fargpt4vclosinggap} & 7B\cite{bai2023qwen} & 1.5-26B\cite{chen2024fargpt4vclosinggap} & 1.5-26B\cite{chen2024fargpt4vclosinggap} & 1.5-26B\cite{chen2024fargpt4vclosinggap} & 1.5-26B\cite{chen2024fargpt4vclosinggap} & 9B\cite{glm2024chatglm} & 1.5-8B\cite{hu2024docowl} & 1.5-8B\cite{hu2024docowl} & 1.5-8B\cite{hu2024docowl} & 1.5-8B\cite{hu2024docowl} \\ 
 Performance  & \underline{60.4} & \underline{50.4 }& 59.0 & 42.0 & 58.4 & \textbf{90.9} & \textbf{83.8} & \underline{72.5} & \textbf{80.6 }& \textbf{77.6} & \underline{40.6}  & \underline{68.8} & \underline{246.4} & \underline{80.2 } \\  

 \midrule \multicolumn{11}{l}{\textit{Closed-source API}} \\
        GPT-4V~\cite{openai2023gpt4} & 43.5 & 49.2 & \underline{59.9} & \textbf{56.0} & --- & \underline{88.4} & 78.5 & 75.1 & 78.0 & 51.6 & --- & --- & --- & ---  \\
        Gemini-Pro~\cite{geminiteam2023gemini} &  --- & --- & \textbf{75.0} & 49.3 & \textbf{70.6} & 88.1 & 74.1 & \textbf{75.2} & 74.6 & 68.0 & --- & --- & --- & ---  \\ 
\midrule 
\rowcolor[HTML]{F2F3F5} 
        IXC-2.5-7B & \textbf{69.1} & \textbf{58.8} & 55.8 & 46.9 & \underline{67.1} &\textbf{ 90.9} & \underline{82.2} & 69.9 & \underline{78.2} & \underline{69.0} & \textbf{53.6 }& \textbf{71.2}  & \textbf{307.5} & \textbf{85.2 }\\

 \bottomrule
\end{tabular} }}
\vspace{-2mm}
\caption{Comparison with closed-source APIs and previous open-source SOTAs \textbf{on Video Benchmarks and Structural High-resolution Benchmarks.} The best results are \textbf{bold} and the second-best results are \underline{underlined}. $^{*1}$ We scale the score from $0\sim 3 $ to $0\sim 100 $ for easier understanding. $^{*2}$ We report the determinism part (MCQA, Y/N, Caption Match) of TempCompass as the evaluation using GPT-3.5 is not stable.}
\vspace{-3mm}
\label{tab:sota_comp_1}
\end{table*}

\begin{table*}[t!]
\footnotesize
\centering
\setlength{\tabcolsep}{0.2mm}{
\resizebox{1.\linewidth}{!}{
\begin{tabular}{l|c|ccccccccccccccc}
\toprule

 & MMDU &  MMStar & RealWQA & MathVista & AI2D & MMMU & MME & MMB & MMB$_{CN}$ & MMB$_{1.1}$ & SEED$^I$ & MM-Vet & HallB \\ 
 
\midrule 
Open-Source  & LLaVa1.6&  InternVL & WeMM & WeMM & InternVL & 360VL & InternVL & InternVL1.5 & InternVL1.5 & InternVL1.5 & WeMM & GLM-4v & WeMM   \\ 
Previous SOTA  & 8B\cite{llava1_5} & 1.5-26B\cite{chen2024fargpt4vclosinggap} & 8B\cite{wemm} & 8B\cite{wemm} & 1.5-26B\cite{chen2024fargpt4vclosinggap} & 70B\cite{360vl} & 1.5-26B\cite{chen2024fargpt4vclosinggap} & 1.5-26B\cite{chen2024fargpt4vclosinggap} & 1.5-26B\cite{chen2024fargpt4vclosinggap} & 1.5-26B\cite{chen2024fargpt4vclosinggap} & 8B\cite{wemm} & 14B\cite{glm2024chatglm} & 8B\cite{wemm}  \\
Performance  & 42.8& \underline{57.1}  & \textbf{68.1} & \underline{54.9} & \underline{80.6} & \underline{53.4} & \underline{2,189.6 }& \textbf{82.3} & \underline{80.7} & \underline{79.7} & \textbf{75.9} & \underline{58.0} & \textbf{47.5} \\  

 \midrule \multicolumn{11}{l}{\textit{Closed-source API}} \\
        GPT-4V~\cite{openai2023gpt4} & 66.3& \underline{57.1}  & \underline{68.0} & 47.8 & 75.5 & \textbf{56.8} & 1,926.5 & 81.3 & 80.2 & \textbf{79.8} & 69.1 & 56.8 & \underline{46.5}  \\ 
        Gemini-Pro~\cite{geminiteam2023gemini} & --- &  42.6 & 64.1 & 45.8 & 70.2 & 47.9 & 1,933.3 & 73.9 & 74.3 & 73.9 & 70.7 & \textbf{59.2} & 45.2  \\ 
\midrule 
\rowcolor[HTML]{F2F3F5} 
        IXC-2.5-7B & 56.6 & \textbf{59.9} & 67.8 & \textbf{63.8} & \textbf{81.5 }& 42.9 & \textbf{2,229.0} &\underline{ 82.2} & \textbf{80.8} & 79.4 &\underline{ 75.4} & 51.7 & 42.4 \\

 \bottomrule
\end{tabular}}}
\vspace{-2mm}
\caption{Comparison with closed-source APIs and previous open-source SOTAs \textbf{on Multi-Turn Multi-Image Dialog and General Visual QA Benchmarks.}  The best results are \textbf{bold} and the second-best results are \underline{underlined}.}
\vspace{-3mm}
\label{tab:sota_comp_2}
\end{table*}

\noindent \textbf{Preference Data Collection.}
We use the fine-tuned model $\pi_{\text{ref}}$ to generate diverse responses for each prompt in the augmented instruction tuning data $\mathcal{D}^{*}$, using different random seeds. This yields a collection of 80,000 prompt-response pairs. Next, we employ the GPT-4o model to label 2,000 responses with chosen or rejected decisions and give the reasons, which serve as our reward modeling data. We then train a reward model $\pi_{\text{rm}}$, sharing the same architecture of $\pi_{\text{ref}}$, on the reward modeling data. The reward model is used to make the chosen or rejected prediction on the remaining prompt-response pairs. These selected responses are then used to construct the pair data $\mathcal{D}^{p} = \{x, y_{w}, y_{l}\}$, while $x$, $y_{w}$ and $y_{l}$ refer to the prompt, chosen response and rejected response, respectively. Ultimately, we obtain a total of 30,000 preference data $\mathcal{D}^{p}$ for DPO~\cite{rafailov2024direct} alignment.

\noindent \textbf{DPO Alignment.}
We use the DPO algorithm to update the SFT model $\pi_{\text{ref}}$ on target policy from the preference data $\mathcal{D}^{p}$:
\begin{equation}\label{eq:dpo}
\begin{aligned}
    \mathcal{L_{\text{DPO}}}(\pi_{\theta}, \pi_{\text{ref}}) = \mathbb{E}_{x, y_{w}, y_{l} \sim \mathcal{D}^{p}} \\
    [- \log \sigma(\beta \log( \frac{\pi_{\theta}(y_{w}|x)}{\pi_{\text{ref}}(y_{w}|x)} ) -\beta \log( \frac{\pi_{\theta}(y_{l}|x)}{\pi_{\text{ref}}(y_{l}|x)} ) )].
\end{aligned}
\end{equation}
In practice, we use LoRA with a rank of $256$ to get the DPO model $\pi_{\theta}$. We observe that our model tends to prioritize minimizing the likelihood of dis-preferred responses $y_{l}$ over maximizing the likelihood of preferred responses $y_{w}$ to avoid generating inappropriate or low-quality content.

In summary, our scalable pipeline consists of three primary components. First, we address the challenge of limited instruction tuning data by re-writing original prompts into augmented prompts. Next, we generate diverse responses using different random seeds, enabling the exploration of various creative possibilities. Finally, we apply the DPO algorithm to the chosen and rejected responses to refine our model's performance. Through our pipeline, our model is capable of generating high-quality articles.
\begin{table*}[t!]
\footnotesize
\centering
\begin{tabular}{l|cccc|c|c}
\toprule

 & Block-Match&  Text  & Position & Color & CLIP & Average \\ 
 
 \midrule 
 \multicolumn{7}{l}{\textit{Closed-source API}} \\
        GPT-4V~\cite{openai2023gpt4} & \textbf{85.8} & \textbf{97.4} & \underline{80.5} & 73.3 & \textbf{86.9} & \underline{84.8} \\ 
        Gemini-Pro~\cite{geminiteam2023gemini} & 80.2 & 94.6 & 72.3 & 66.2 & 83.9 & 79.4 \\ 
\midrule 
\multicolumn{7}{l}{Open-source} \\
        WebSight VLM-8B~\cite{laurenccon2024unlocking} & 55.9 & 86.6 & 77.3 & \underline{79.4} & \underline{86.5} & 77.1 \\ 
        CogAgent-Chat-18B~\cite{hong2024cogagent} & 7.1 & 18.1  &13.3 & 13.0 & 75.5 & 25.4 \\ 
        Design2Code-18B~\cite{si2024design2code}  & 78.5 & \underline{96.4} & 74.3 & 67.0 & 85.8 & 80.4 \\ 
\midrule 
\rowcolor[HTML]{F2F3F5} 
        IXC-2.5-7B & \underline{81.9} & 95.6 & \textbf{80.9} & \textbf{80.8} & \underline{86.5} & \textbf{85.1} \\ 
 
 \bottomrule
\end{tabular}
\vspace{-2mm}
\caption{Screenshot-to-code. Comparison with closed-source APIs and open-source models on \textbf{Design2Code benchmark}. The best results are \textbf{bold} and the second-best results are \underline{underlined}.}
\vspace{-3mm}
\label{tab:sota_comp_design2code}
\end{table*}

\section{Experiments}\label{sec:experiments}
In this section, we validate the benchmark performance of our InternLM-XComposer-2.5 (IXC-2.5) after supervised fine-tuning. 

\subsection{LVLM Benchmark results.}
In Table \ref{tab:sota_comp_1} and Table \ref{tab:sota_comp_2}, we compare our IXC-2.5 on a list of benchmarks with both closed-source APIs and SOTA open-source LVLMs (with comparable model size). Here we report video understanding results on MVBench~\cite{li2024mvbench}, MLVU~\cite{zhou2024mlvu}, MME-Video~\cite{fu2024video}, MMBench-Video~\cite{fang2024mmbench}, TempCompass~\cite{liu2024tempcompass}. 
For Structural High-resolution understanding, we report results on  DocVQA~\cite{docvqa}, ChartQA~\cite{masry2022chartqa}, InfographicVQA~\cite{infovqa}, TextVQA~\cite{textvqa}, OCRBench~\cite{ocrbench}, DeepForm~\cite{deepform}, WikiTableQuestion~(WTQ)~\cite{pasupat-liang-2015-compositional}, Visual MRC~\cite{tanaka2021visualmrc}, and TabFact~\cite{chen2020tabfact}.
For general visual question answering, we report results on MMStar~\cite{mmstar}, RealWorldQA\cite{RealWorldQA},
MathVista~\cite{lu2024mathvista}, MMMU~\cite{yue2023mmmu}, AI2D~\cite{kembhavi2016diagram}, MME~\cite{fu2023mme}, MMBench (MMB) ~\cite{MMBench}, MMBench-Chinese (MMB$^{CN}$)~\cite{MMBench}, MMBench-v1.1~(MMB$^{v1.1}$)~\cite{MMBench}, SEED-Bench Image Part~(SEED$^{I}$)\cite{li2023seedbench}, MM-Vet~\cite{yu2023mmvet}, HallusionBench (HallB)~\cite{guan2023hallusionbench}. For Multi-True Multi-Image dialogue, we evaluate IXC-2.5 on MMDU~\cite{liu2024mmdu} benchmark. For webpage crafting, we report a subtask screenshot-to-code~\cite{si2024design2code} since benchmarks for others are not available in the community.

The evaluation is mainly conducted on the OpenCompass VLMEvalKit~\cite{2023opencompass} for the unified reproduction of the results.

% \textbf{{\color{c1}{Video Benchmarks}}}, \textbf{{\color{c2}{Structural High-resolution Benchmarks}}}, \textbf{{\color{c3}{General Visual QA Benchmarks}}}, \textbf{{\color{c4}{Multi-True Multi-Image Benchmark}}}, and \textbf{{\color{c5}{Webpage Crafting Benchmark}}}.

\noindent\textbf{Comparison on Video Understanding Benchmarks.}
As demonstrated in Table \ref{tab:sota_comp_1}, IXC-2.5 exhibits competitive performance on fine-grained video understanding tasks, outperforming open-source models on 4 of the 5 benchmarks and being on par with Closed-Source APIs. For example, IXC-2.5 reaches 69.1 on the MVBench, $+8.7\%$ higher than the previous SOTA method VideoChat2-7B and outperforms GPT-4V with $+25.6\%$. For the recent challenging MMBench-Video, IXC-2.5 reaches the SOTA performance on open-source models and performs close to Gemini-Pro. 

\noindent\textbf{Comparison on Structural High-resolution Benchmarks.}
\noindent Benefiting from the unified image partition strategy, IXC-2.5 could handle diverse kinds of images. Table \ref{tab:sota_comp_1} reports its performance on several structural high-resolution benchmarks. IXC-2.5 with only 7B parameters performs on par with the current large open-source LVLMs and close-source APIs. For example, IXC-2.5 gets $90.9\%$ on the DocVQA test set, the same as InternVL-1.5 which has nearly $4\times$ parameters. For the highly structured form and table understanding tasks, IXC-2.5 outperforms DocOwl 1.5-8B with  $+13.0\%$, $+2.4\%$, $+5.0\%$ on WikiTableQuestion, DeepForm and TableFace respectively.

\noindent\textbf{Comparison on Multi-Image Multi-Turn Benchmarks.}
\noindent IXC-2.5 is capable of taking multiple images as input and conducting multi-round free-form dialogue based on them. We evaluate it quantitatively on the newly proposed MMDU benchmark~\cite{liu2024mmdu}. As shown in Table \ref{tab:sota_comp_2}, the IXC-2.5 model demonstrates superior performance, outperforming the previous SOTA open-source model by a significant margin of 13.8\%. This notable improvement highlights the effectiveness of our approach in advancing the capabilities of multi-image and multi-turn understanding.

\noindent\textbf{Comparison on General Visual QA Benchmarks.}
\noindent IXC-2.5 is designed as a general LVLM to handle diverse multi-modal tasks. Here we report its performance on general visual QA benchmarks. As shown in Table \ref{tab:sota_comp_2}, the IXC-2.5 shows superb performance on these benchmarks and on par with current large open-source LVLMs and closed-source APIs. For example, IXC-2.5 gets $59.9\%$ on the challenging MMStar and outperforms GPT-4V and Gemini-Pro. On the RealWorldQA, IXC-2.5 also performs better than Gemini-Pro and close to GPT-4V.

\noindent \textbf{Comparison on Screenshot-to-code Benchmark.} Table~\ref{tab:sota_comp_design2code} presents the comparison results on the Design2Code~\cite{si2024design2code} benchmark that assesses the ability to translate visual design into code implementation. Our IXC-2.5 even surpasses the GPT-4v on average performance, which highlights the potential of IXC-2.5 to excel in bridging the gap between visual design and code implementation.

\section{Conclusion}
We have introduced InternLM-XComposer-2.5 (IXC-2.5), a cutting-edge Large Vision-Language Model (LVLM) boasting long-contextual input and output capabilities that enable advanced features such as ultra-high resolution image understanding, fine-grained video understanding, multi-turn multi-image dialogue, webpage generation, and article composing.
Our comprehensive experiments demonstrate that IXC-2.5 achieves competitive performance, remarkably, with a relatively modest 7B Large Language Model (LLM) backend.

Our model sets out a promising research direction that can extend to a more contextual multi-modal environment, including long-context video understanding (\eg, long movies) and long-context interaction history, to better assist humans in real-world applications. 

\noindent\textbf{Acknowledgements:} We deeply express our gratitude to Prof. Chao Zhang from Tsinghua University for suggestions about audio models and tools.
\clearpage
{
    \small
    \bibliographystyle{ieeenat_fullname}
    \bibliography{main}
}


\end{document}